\documentclass[]{mva_style}
\usepackage{graphicx}
\usepackage{times}
\usepackage{epsfig}
\usepackage{graphicx}
\usepackage{amsmath}
\usepackage{amssymb}
\usepackage{booktabs}
\usepackage{verbatim}
\usepackage{multirow}
\usepackage{subfigure}
\usepackage{comment}
\usepackage[font=small,textfont=it,labelfont=bf]{caption}
\usepackage{xcolor}

\finalcopy 

\newcommand{\ra}[1]{\renewcommand{\arraystretch}{#1}}

\begin{document}
\title{Age Prediction From Face Images Via Contrastive Learning}

\author{Yeongnam Chae $\qquad \qquad$ Poulami Raha $\qquad \qquad$ Mijung Kim $\qquad \qquad$  Bj\"orn Stenger \vspace{1mm} \\ 
Rakuten Institute of Technology, Rakuten Group, Inc.\\
{\tt\small \{yeongnam.chae, poulami.raha, mijung.a.kim, bjorn.stenger\}@rakuten.com}
}


\maketitle

\section*{\centering Abstract}
\textit{
This paper presents a novel approach for accurately estimating age from face images, which overcomes the challenge of collecting a large dataset of individuals with the same identity at different ages. Instead, we leverage readily available face datasets of different people at different ages and aim to extract age-related features using contrastive learning. Our method emphasizes these relevant features while suppressing identity-related features using a combination of cosine similarity and triplet margin losses.
We demonstrate the effectiveness of our proposed approach by achieving state-of-the-art performance on two public datasets, FG-NET and MORPH~II.}

\section{Introduction}

 Age estimation from facial images has found application in various fields. However, the different attributes in a face image present distinct visual features that can be challenging to disentangle. While facial feature extractors have been pre-trained for specific tasks such as face recognition, they are not well-suited for tasks like age estimation. Collecting a large dataset of face images of the same individuals at different ages is more difficult than collecting a dataset of different individuals. Therefore, the key question is how to develop a method for learning aging-related features that are not influenced by  identity-related features.

Face age estimation methods based on convolutional neural networks have made significant progress and can be grouped into three categories: classification~\cite{dehshibi2010new,1262530,ramesha2010feature,10.1007/978-3-540-74549-5_49}, regression~\cite{DORNAIKA2020112942,4523958,4531189,5298825,or-cnn} and ranking~\cite{7001258,5597533,ohrank,ranking-cnn} approaches. Recently, self-supervised ~\cite{occo} and attention-based approaches~\cite{adpf} have been proposed. 
However, most of these techniques have relied on information from individual face images, causing the model to be biased towards features associated with attributes like identity, which hinders the model from focusing on relevant but sparse features related to age, such as small wrinkles or skin texture.
Moreover, some studies have explored comparative approaches~\cite{CRCNN,svrt,LSDML,RCL}, which aim to learn ranking or transformation information based on relative age differences. In contrast, the proposed approach mainly focuses on how to emphasize sparser features related to age while penalizing non-relevant features related to identity through contrasting the features extracted from images from the same age group.

\begin{figure}[t]
\begin{center}
\includegraphics[width=0.99\linewidth]{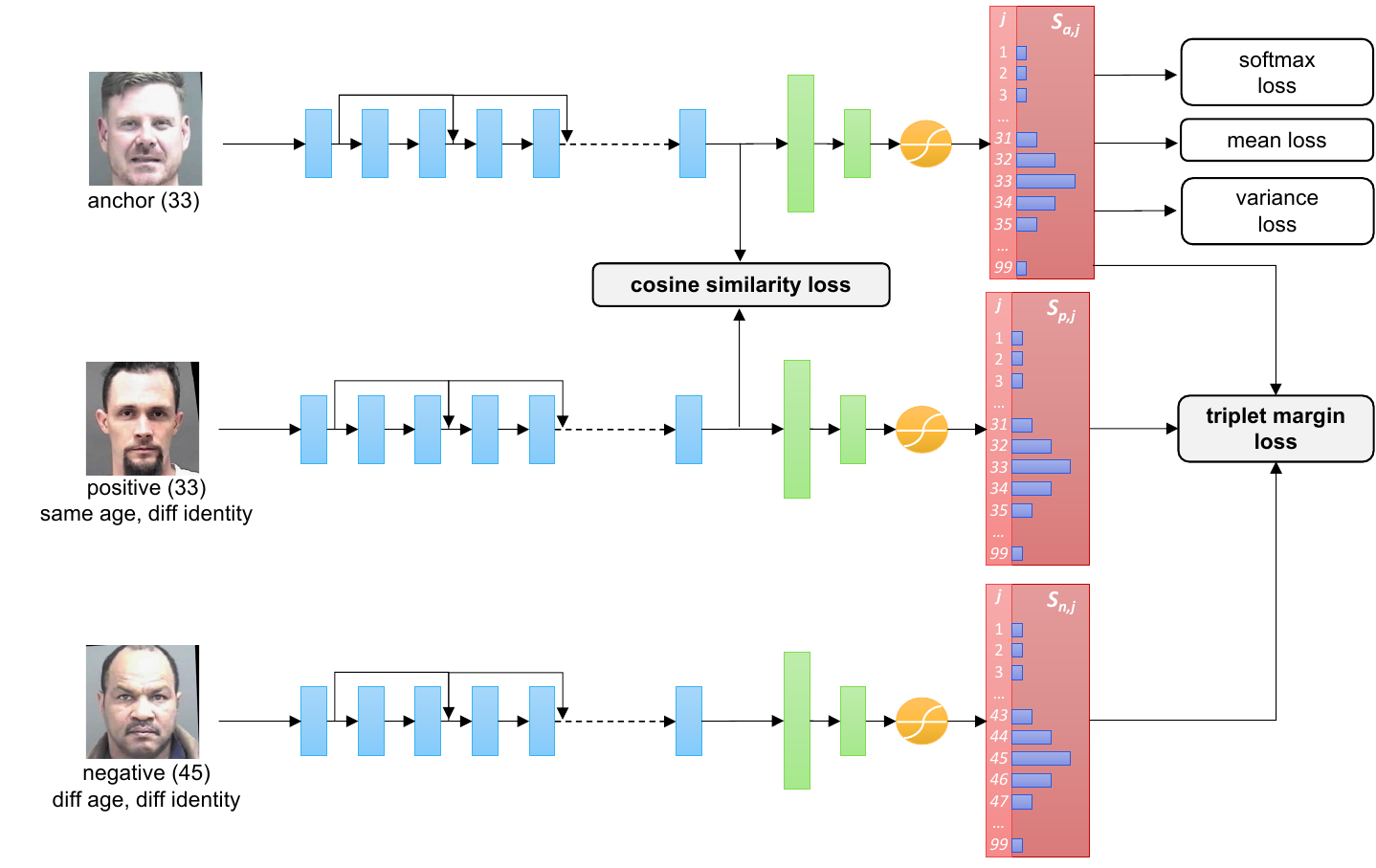}
\end{center}
\caption{{\bf Contrastive Learning for Age Estimation} using cosine similarity loss for positive samples with the same age as the anchor image and triplet margin loss for negative samples with a different age.}
\label{fig:overall}
\end{figure}

We present a new approach for face age estimation that leverages contrastive learning. Our method aims to suppress identity-related features while emphasizing age-related ones. To learn identity-independent age features, we use triplets of images. Given an anchor face image, we sample two images - one of the same age (positive sample) but with a different identity, and one of a different age (negative sample). By comparing the anchor image to both the positive and negative samples, we jointly minimize a cosine similarity loss and a triplet margin loss, see Figure~\ref{fig:overall}.
Owing to the large number of possible triplet samples, the method is data-efficient and able to learn age prediction from small datasets without resorting to additional data sources, as shown in the experiments on public datasets.




\section{Method}




We use contrastive learning to suppress identity-related features in order to compare face images with the same age but different identity. We extract the facial feature vector from an anchor image, ${\bf x}_a$, and a positive sample, ${\bf x}_p$.
To ensure that the positive sample ${\bf x}_p$ of the given anchor ${\bf x}_a$ is free from identity-related features, we select it from a set of face images ${\cal P}({\bf x}_a)$ that have the same age but different identity labels:
\begin{equation}
\label{eq:positive}
{\cal P}({\bf x}_a) = \{{\bf x}_p ~ | ~ y_p = y_a \; \mathrm{and} \; i_p \neq i_a\} ,
\end{equation}
where $i_a$ and $i_p$ are the identity labels and $y_a$ and $y_p \in \{1,2,..., A\}$ are age labels of the anchor and the positive sample, respectively. $A$ is the maximum age label in a dataset.
We extract ResNet-18 features ${\bf f}_a = R({\bf x}_a)$, ${\bf f}_p = R({\bf x}_p)$, where ${\bf f} \in \mathbb{R}^{D}$ are features before the last fully connected (FC) layer. $D=512$ is the dimension of the feature vector. We compute the cosine similarity loss $L_{c}$~\cite{cosine} between features ${\bf f}_a$ and ${\bf f}_p$:
\begin{equation}
\label{eq:cos}
  L_{c}(\mathbf{f}_a, \mathbf{f}_p) = \frac{{\mathbf{f}_a}^T \mathbf{f}_p}{ \| \mathbf{f}_a \|  \| \mathbf{f}_p \|}  .
\end{equation}
The cosine distance has been widely used to compare face images or features~\cite{ deng2019arcface,huang2020curricularface,kim2022adaface,liu2017sphereface,meng2021magface,wang2018cosface}. 

The probabilities computed in the final softmax output of the network for the anchor image define a distribution over age values.
%
Let $s_{a,j}$ denote the probability that sample ${\bf x}_a$ has age label~$j$.
We compute a mean loss, $L_{m}$, and a variance loss, $L_{v}$~\cite{mv}:
\begin{equation}
\label{eq:mean}
  L_{m}(\mathbf{s}_a) = \sum_{j=1}^{A} {j s_{a,j} } - y_a \; ,
\end{equation}
\begin{equation}
\label{eq:var}
  L_{v}(\mathbf{s}_a) = \sum_{j=1}^{A} {s_{a,j}}  \left(j - \sum_{k=1}^{A}{k \, s_{a,k}}\right)^2 \; .
\end{equation}


\paragraph{Ternary contrast via triplet margin loss.}
To use contrastive learning for different ages, we adopt a triplet margin loss~\cite{facenet}. The triplet margin loss is a loss function where the anchor image is compared with a positive and a negative sample. The distance from the anchor to the positive sample is minimized, while the distance to the negative sample is maximized. We use a negative sample ${\bf x}_n$ with a different age and identity to contrast with the positive sample and the anchor.
\begin{equation}
\label{eq:negative}
{\cal N}({\bf x}_a) = \{{\bf x}_n ~ | ~ y_n \neq y_a \; \textrm{and} \; i_n \neq i_a\} 
\end{equation}
The triplet loss is applied to the softmax probability, $\mathbf{s}$ as follows:
\begin{equation}
\label{eq:triplet}
  L_{t}(\mathbf{s}_a, \mathbf{s}_p, \mathbf{s}_n) = \max (\|\mathbf{s}_a - \mathbf{s}_p\|^2 - \|\mathbf{s}_a - \mathbf{s}_n\|^2 + \alpha, 0) ,
\end{equation}
where $\mathbf{s}_n$ is softmax probability of the negative image and $\alpha$ is a margin between positive and negative samples. 
 The overall loss function is defined as:
\begin{equation}
    \label{eq:loss}
    L_{\textrm{total}} = L_s + \lambda_m L_m + \lambda_v  L_v + \lambda_c  L_c + \lambda_t L_t \; ,
\end{equation}

where $L_s$ denotes the softmax loss, $\lambda$ denote hyper-parameters that balance the influence of each loss term. We empirically set $\lambda_m$ to 0.2 and $\lambda_v$ to 0.05.  We experimentally evaluate the values of the binary and ternary loss terms, $\lambda_c$ and $\lambda_t$.

\section{Experiments}


\paragraph{Datasets \& protocols.}
The {\it MORPH II} dataset is a face dataset, containing 55,134 images of 13,618 individuals. Ages range from 16 to 77 with a median age of 33. To be consistent with prior work, the five-fold random split (RS) and five-fold subject exclusive (SE) protocols are used in the experiments~\cite{morph}.

The {\it FG-NET} dataset contains 1,002 face images from 82 individuals with ages ranging from 0 to 69 years~\cite{fg-net}. We evaluate using the commonly used leave-one-person-out (LOPO) protocol. Table~\ref{table:age-range} shows the age distributions in the MORPH II and FG-NET datasets.

\setlength{\tabcolsep}{4pt}
\begin{table}[t]
\begin{center}
\caption{{\bf Age distributions} in MORPH II and FG-NET.}
\label{table:age-range}
\begin{tabular}{lrrrr}
\hline\noalign{\smallskip}
Dataset  & 0-19 & 20-39 & 40-59 &  $>$60 \\
\noalign{\smallskip}
\hline
\noalign{\smallskip}
MORPH II & 7,469 & 31,682 & 15,649 & 334\\
FG-NET & 710 & 223 & 61 & 8\\
\hline
\end{tabular}
\end{center}

\end{table}
\setlength{\tabcolsep}{1pt}

\paragraph{Implementation details \& evaluation metric.}
We first perform facial alignment of all images using five landmarks detected with MTCNN \cite{mtcnn}. These aligned face images are then normalized to a size of $256 \times 256 \times 3$. Subsequently, we extract features from the normalized face images by utilizing a ResNet-18 model that has been pre-trained on ImageNet.
We apply a series of augmentation techniques, including random affine transformations with slight variations, random vertical flips, and random crops to $224 \times 224$ during the training phase. We optimize the model parameters using the Adam optimizer with an initial learning rate of 0.001 and train for 100 epochs, using a batch size of 64.
As error metric we use the mean absolute error (MAE), which is defined as the L1 distance between the predicted age, $\widehat{y}_i$, of image ${\bf x}_i$ and its ground-truth age, $y_i$:
\begin{equation}
\label{eq:mae}
  \mathrm{MAE} = \frac{1}{N}\sum_{i=1}^{N}{| \widehat{y}_i - y_i|}  \; .
\end{equation}

\setlength{\tabcolsep}{4pt}
\begin{table}[t]
\begin{center}
\caption{{\bf Evaluation on MORPH II}.
Mean Absolute Error (MAE) comparison following the RS evaluation protocol. RS(I) indicates pre-training on ImageNet, RS(F) on other large face datasets, including IMDB-WIKI~\cite{dex}, MSCeleb(\dag)~\cite{msceleb}, and FaceAugmentation(\ddag)~\cite{FAdataset}. Parameters: (binary) $\lambda_c = 10$, $\lambda_t = 0$ (ternary) $\lambda_c$ = 10, $\lambda_t = 1$. }
\label{table:sota-morph-rs}
\begin{tabular}{lccccc}
\toprule
Method & RS(I) & RS(F) & Year  \\
\midrule
CRCNN\cite{CRCNN} & 3.74 & &2016\\
OR-CNN\cite{or-cnn} & 3.27 & &2016\\
DEX\cite{dex} & 3.25 & 2.68 &2016\\
ODFL\cite{odfl} & 3.12 & &2017\\
ARN\cite{arn} & 3.00 & &2017\\
Ranking-CNN\cite{ranking-cnn} & 2.96 & & 2017\\
AP\cite{ap} & 2.87 &  2.52 &2017\\
M-LSDML\cite{LSDML}& & 2.89$^\ddag$ &2018 \\
RCL\cite{RCL}& & 2.46 &2018 \\
SVRT\cite{svrt} & & 2.38$^\dag$ &2018\\
MV\cite{mv} & 2.41 & 2.16 &2018\\
C3AE\cite{c3ae} & 2.78 & 2.75 &2019\\
BridgeNet\cite{bridgenet} &  &  2.38 &2019\\
ADVL\cite{advl} &  &  1.94 &2020\\
NRLD\cite{nrld} & 2.35 & {\bf 1.81} &2020\\
OCCO\cite{occo} & 2.29 & &2021\\
ADPF\cite{adpf} & 2.54 & &2022\\
{\bf Ours (binary loss)} & {\bf 2.14} & & \\
{\bf Ours (ternary loss)} & 2.20 & &\\
\bottomrule
\end{tabular}
\end{center}
\end{table}

\setlength{\tabcolsep}{1.2pt}
\setlength{\tabcolsep}{4pt}
\begin{table}[t]
\begin{center}
\caption{
{\bf Evaluation on MORPH II}.
Mean Absolute Error (MAE) comparison following the SE evaluation protocol.}
\label{table:sota-morph-se}
\begin{tabular}{lccc}
\toprule
Methods & SE(I) & SE(F) & Year \\
\midrule
DIF\cite{dif} & 3.00 & &2018\\
RCL\cite{RCL}& & 2.88 &2018 \\
SVRT\cite{svrt}& & 2.87$^\dag$ &2018 \\
MV\cite{mv} & 2.80 & {\bf 2.79} &2018\\
{\bf Ours (binary loss)} & 2.43 & \\
{\bf Ours (ternary loss)} & {\bf 2.37} & \\
\bottomrule
\end{tabular}
\end{center}
\end{table}
\setlength{\tabcolsep}{1.2pt}

\setlength{\tabcolsep}{4pt}
\begin{table}[t]
\begin{center}
\caption{{\bf Evaluation on FG-NET} in terms of Mean Absolute Error (MAE) following the LOPO evaluation protocal. LOPO(I) indicates pre-training on ImageNet, LOPO(F) on the large IMDB-WIKI~\cite{dex} dataset. Parameters: (binary) $\lambda_c = 1$, $\lambda_t = 0$ (ternary) $\lambda_c$ = 1, $\lambda_t = 1$}
\label{table:sota-fgnet}
\begin{tabular}{lccc}
\toprule
Methods & LOPO(I) & LOPO(F) & Year \\
\midrule
DEX\cite{dex} & 4.63 & 3.09 & 2016\\
CRCNN\cite{CRCNN} & 4.13 & & 2016\\
RCL\cite{RCL} & & 4.21 & 2018\\
MV\cite{mv} & 4.10 & 2.68 & 2018\\
C3AE\cite{c3ae} & 4.09 & 2.95 & 2019\\
BridgeNet\cite{bridgenet} &  & 2.56 & 2019\\
NRLD\cite{nrld} & 3.23 & 2.55 & 2020\\
AVDL\cite{advl} &  & {\bf 2.32} & 2020\\
ADPF\cite{adpf} & 2.86 & & 2022\\
{\bf Ours (binary loss)} & 2.42 &\\
{\bf Ours (ternary loss)} & {\bf 2.31} &\\
\bottomrule
\end{tabular}
\end{center}
\end{table}
\setlength{\tabcolsep}{1.2pt}



\begin{figure}[t]
\subfigure[FG-NET] {
\includegraphics[width=0.43\columnwidth]{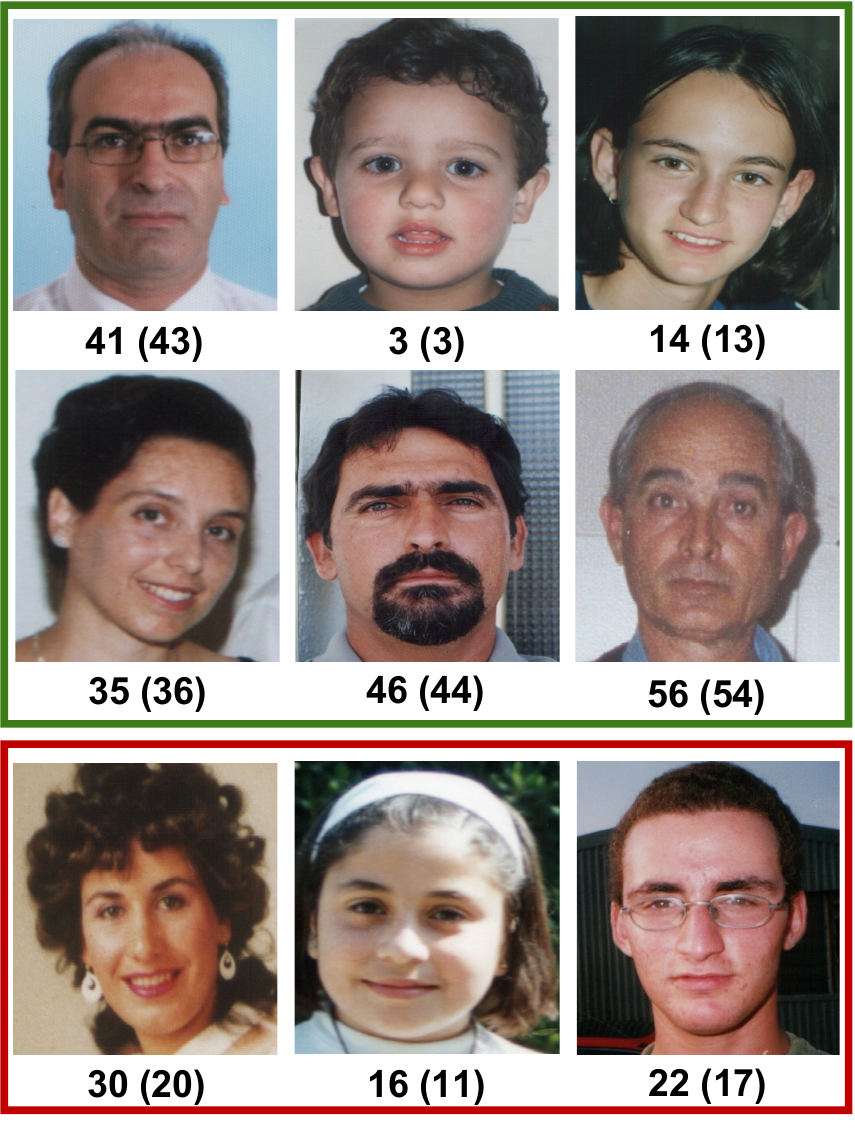} 
} 
\subfigure[MORPH II] {    
\includegraphics[width=0.43\columnwidth]{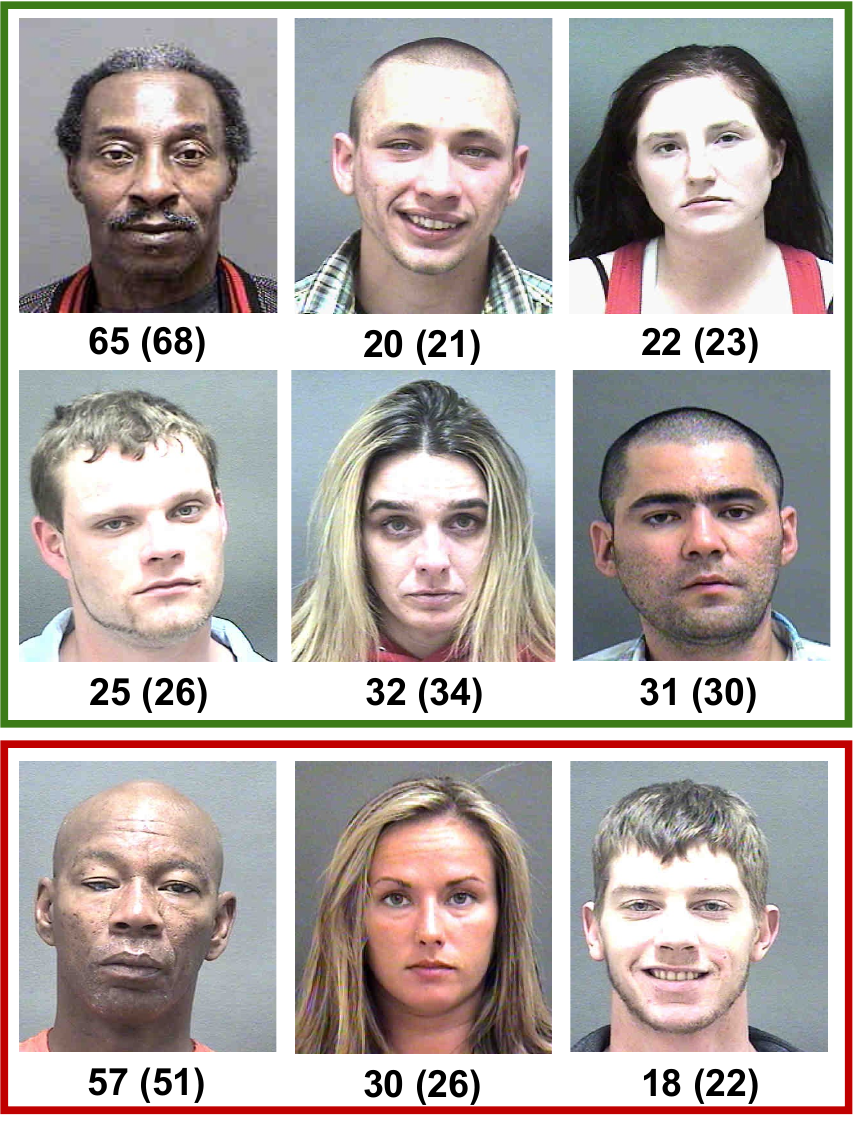} 
}    

\subfigure[Grad-CAM Visualizations] {     
\includegraphics[width=0.90\linewidth]{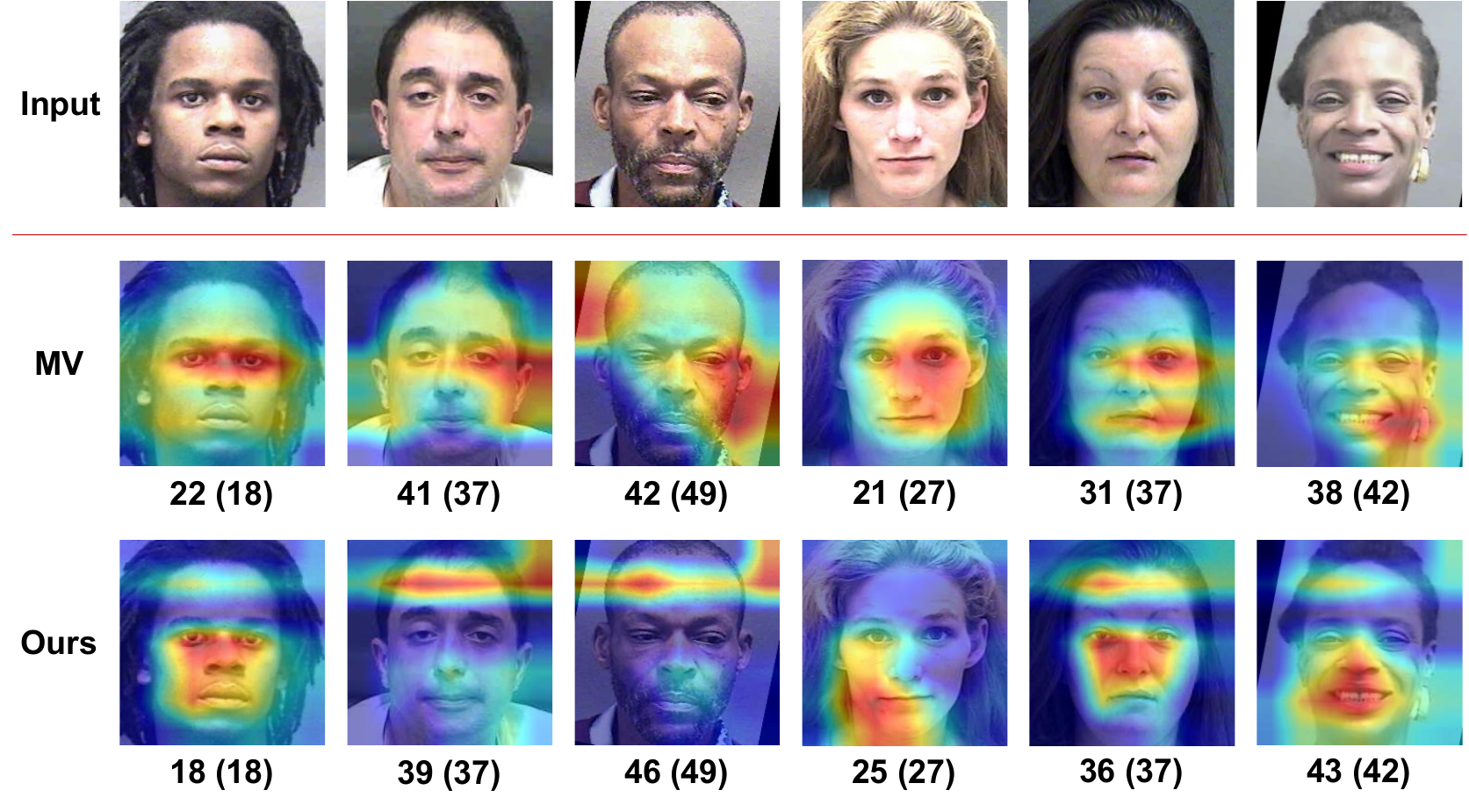} 
}  

\caption{\label{fig:examples} {\bf Example age prediction results} on (a) FG-NET and (b) MORPH II.
Numbers below each image show predicted and ground truth age.
Top two rows show accurate estimation results, the bottom row shows poor estimates. 
(c) A Grad-CAM analysis on MOPRH II samples shows that our method better focuses on features extracted from the face region compared to the Mean Variance method~\cite{mv}. }
\end{figure}

\paragraph{Results on MORPH II.} Table~\ref{table:sota-morph-rs} shows the results following the RS protocol. The proposed model achieves an MAE of 2.14 when using binary contrast ($\lambda_c = 10$, $\lambda_t = 0$). It performs best among all approaches that do not use external data for pre-training. Several approaches employ model pre-trained on large face datasets like IMDB-WIKI~\cite{dex}, MS Celeb 1M~\cite{msceleb}, or FaceAugmentation~\cite{FAdataset}. 
Results following the SE evaluation protocol on the MORPH II dataset are shown in Table~\ref{table:sota-morph-se}. The proposed model achieves an MAE of 2.37 when using ternary contrast ($\lambda_c$ = 10, $\lambda_t = 1$). In the SE protocol, images of individuals who appear in the training set are excluded from the test set.  We observe that in this case the ternary loss improves the performance. 
The coefficients of the individual loss terms are discussed in the ablation study.

\paragraph{Results on FG-NET.} As shown in Table~\ref{table:sota-fgnet}, our method shows good performance on the FG-NET dataset. We observe that our model shows competitive performance with models that use an external dataset, IMDB-WIKI~\cite{dex}. 
By sampling triplets of face images, the proposed method achieves excellent performance on FG-NET without any additional data.

Examples of age estimation results on FG-NET and MORPH II are shown in Fig.~\ref{fig:examples}. The proposed method performs robustly for various age ranges. Poor estimates are typically caused by poor image quality.

\paragraph{Identity invariance.}
To measure the dependence on face identiy, we compare the feature variance when keeping the identity fixed.
A low variance indicates a larger dependency on identiy and vice versa.
We calculate the mean variance of the extracted feature $\mathbf{f}$ and $\mathbf{s}$ by identity and compare it with the Mean Variance (MV) method~\cite{mv} on FG-NET and MORPH II datasets. 
As shown in Table~\ref{table:mvar}, the feature extracted by the proposed method show higher variance than the MV method for the same identity.

\setlength{\tabcolsep}{4pt}
\begin{table}[t]
\begin{center}
\caption{\bf Mean variance for fixed identity.}
\label{table:mvar}
\begin{tabular}{lcrr}
\hline\noalign{\smallskip}
Dataset & Methods & MV & Ours  \\
\noalign{\smallskip}
\hline
\noalign{\smallskip}
FG-NET & $\mu(V_\mathbf{f})$ & 0.18 & {\bf 6.16} \\
FG-NET & $\mu(V_\mathbf{s})$ & 36.03 & {\bf 43.71} \\
MORPH II & $\mu(V_\mathbf{f})$ & 0.18 & {\bf 7.22} \\
MORPH II & $\mu(V_\mathbf{s})$ & 35.87 & {\bf 44.44} \\
\hline
\end{tabular}
\end{center}
\end{table}
\setlength{\tabcolsep}{1.4pt}

 For the qualitative study, a Grad-CAM \cite{gradcam} comparison between the MV method~\cite{mv} and the proposed method on the MORPH II dataset is shown in Fig.~\ref{fig:examples}. MV focuses on the entire face area, whereas the proposed model concentrates mainly on forehead regions for people in their 30s and 40s. In case of teens and 20s, it focuses on areas around the nose and mouth. The eye related features like eye shape, eye size and eyebrow shape are reported as distinctive feature for face identification than other facial features~\cite{abudarham2016reverse}. On the other hand,  features related to wrinkles around the forehead, nose and mouth region are important features for age estimation~\cite{ng2015will}. As shown in Fig.~\ref{fig:examples}, the proposed method makes model less dependent on the identity related features and emphasize the features related to the age.


\paragraph{Ablation study on loss functions.}
To evaluate the combination of different loss functions we study binary and ternary contrast by including or excluding the triplet margin loss. In addition, we compare cosine similarity with a Kullback-Leibler (KL) divergence loss. Cosine similarity is a common to measure of feature similarity while the KLD loss is used for continuous distributions. We adopt the KLD loss on the softmax probabilities for anchor and positive samples:
\begin{equation}
\label{eq:kld}
  L_{\mathrm{KLD}}(\mathbf{s}_a, \mathbf{s}_p) = \frac{1}{A}\sum_{j=1}^{A}{\mathbf{s}_{p,j} (\log{\mathbf{s}_{p,j}} - \mathbf{s}_{a,j})} \; .
\end{equation}
The MAE results for different losses are shown in Table~\ref{table:ablation-structure}. For both binary and ternary contrast, the KLD loss performs worse than the cosine similarity loss. The triplet margin loss improves the performance on small datasets such as FG-NET. The cosine similarity loss improves the accuracy when there is sufficient training data such as in the MORPH II dataset. Overall, cosine similarity loss with triplet margin loss shows the best performance for both small and large datasets. 

\begin{figure}[t]
\includegraphics[width=\columnwidth]{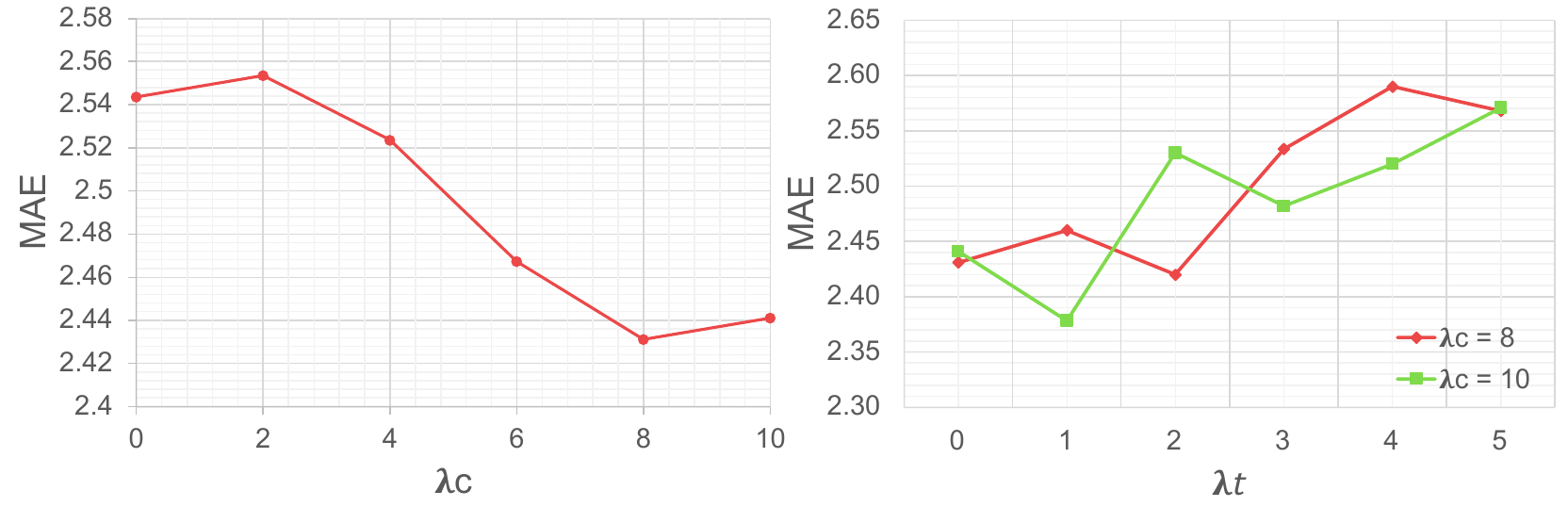}
\caption{{\bf Ablation on loss weights.} MAE on MORPH II (SE) for varying $\lambda_c$ and $\lambda_t$. (left)  varying $\lambda_c$ with  $\lambda_t = 0$, (right) varying $\lambda_t$ with $\lambda_c = 8$ or $ 10$.}
\label{fig:lambda} 
\end{figure}

\paragraph{Loss coefficients.}
In order to select weights for the individual loss terms, we evaluate different weight combinations. We fix coefficients $\lambda_m$, $\lambda_v$ for the mean and variance terms to 0.2 and 0.05, respectively. We change the coefficient of the cosine similarity loss, $\lambda_c$ without negative channel and the triplet margin loss (i.e. $\lambda_t = 0$). We change it from 0 to 10 because an absolute scale of the cosine similarity loss is ten times smaller than either mean loss or variance loss.  We measure the MAE with the SE protocol on the MORPH II dataset, see Fig. \ref{fig:lambda} (a). We observe that cosine similarity  shows good performance for values $\lambda_c=8$ and $\lambda_c=10 $.
We vary the triplet loss coefficient, $\lambda_t$ from 0 to 5. We measure the MAE on MORPH II (SE protocol). As shown in Fig.~\ref{fig:lambda} (b), the cosine similarity loss is a poor choice ($\lambda_t = 0$). Compared with the triplet margin loss, the cosine similarity loss has a relatively larger influence on the performance of the model. The minimum MAE is obtained for $\lambda_t = 1$ and $\lambda_c=10$.  

\setlength{\tabcolsep}{4pt}
\begin{table}[t]
\begin{center}
\caption{{\bf Ablation study.} MAE for different loss terms. Best and second-best results are shown in bold and underlined.}
\label{table:ablation-structure}
 \ra{1.2}\resizebox{1.0\linewidth}{!}{
\begin{tabular}{lccc}
\hline\noalign{\smallskip}
\multirow{2}{*}{Loss} & \small FG-NET & \small MORPH II & \small MORPH II \\
& \small (LOPO) & \small (RS) & \small (SE) \\
\noalign{\smallskip}
\hline
\noalign{\smallskip}
MV & 4.10 & 2.41 & 2.80 \\
MV + KLD & 2.43 & 2.34 & 2.85\\
MV + Cosine & 2.42 & {\bf 2.19} & \underline{2.54}\\
MV + Triplet & {\bf 2.30} & 2.49 & 2.88\\
MV + KLD + Triplet & 2.33 & 3.13 & 2.85\\
MV + Cosine + Triplet & \underline{2.31} & \underline{2.24} & {\bf 2.50}\\
\hline
\end{tabular}}
\end{center}
\end{table}
\setlength{\tabcolsep}{1.4pt}

\section{Conclusion}

We introduced a method for age estimation from face images via contrastive learning from triplets of face images. Our proposed approach focuses on sparser features that are more relevant to age by penalizing non-relevant features that are associated with identity. To encourage the similarity of positive samples, we leverage cosine similarity, and we employ a ternary loss to increase the distance to negative samples. 
Experiments on the MORPH II and FG-NET datasets demonstrated the effectiveness of our proposed method, which achieved state-of-the-art results.

{\small
\bibliographystyle{ieee_fullname}
\sloppy
\bibliography{face_age}
}

\end{document}